\documentclass[10pt,twocolumn,letterpaper]{article}

\usepackage[pagenumbers]{cvpr} 

\usepackage[dvipsnames]{xcolor}

\definecolor{cvprblue}{rgb}{0.21,0.49,0.74}
\usepackage[pagebackref,breaklinks,colorlinks,citecolor=cvprblue]{hyperref}
\usepackage{mmstyle}
\usepackage{amsmath}
\usepackage{pifont}
\usepackage[accsupp]{axessibility}
\newlength\savewidth\newcommand\shline{\noalign{\global\savewidth\arrayrulewidth
		\global\arrayrulewidth 1pt}\hline\noalign{\global\arrayrulewidth\savewidth}}
\def\paperID{1310} 
\def\confName{CVPR}
\def\confYear{2024}

\title{PACER+: On-Demand Pedestrian Animation Controller in Driving Scenarios}

\author{Jingbo Wang$^{1*}$ ~ Zhengyi Luo$^{2*}$  ~ Ye Yuan$^3$ ~ Yixuan Li$^4$ ~ Bo Dai$^1$ \\
\normalsize $^1$Shanghai AI Lab ~$^2$Carnegie Mellon University ~ $^3$NVIDIA ~ $^4$The Chinese University of Hong Kong \\
\normalsize * Donates Equal Contribution}

\begin{document}
\twocolumn[
	{
	\renewcommand\twocolumn[1][]{#1}%
	\maketitle
	\begin{center}
	\includegraphics[width=1.0\linewidth]{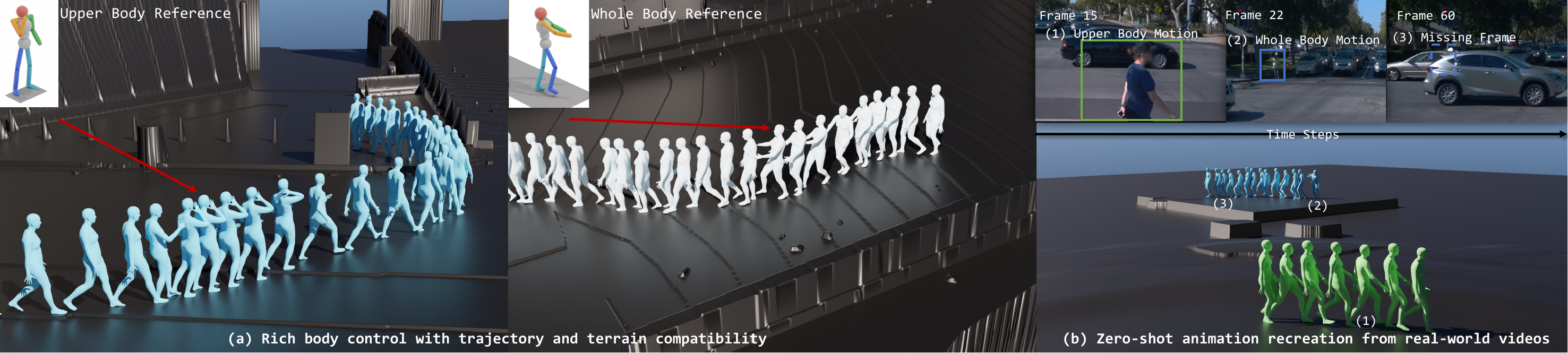}
	\captionof{figure}{\small We showcase the effectiveness of our proposed framework in synthetic and real-world driving scenarios. Our framework excels at generating physically realistic animations that adhere to provided trajectories while offering extensive control over the upper and full body movements. Additionally, our framework demonstrates the remarkable ability to recreate pedestrian animations with occlusions from real-world videos in a \textit{zero-shot} manner. These inherent capabilities make our framework a robust and versatile approach for\textbf{on-demand} pedestrian animation in driving scenarios.}		
	\label{fig:teaser}
	\end{center}
	}]

\begin{abstract}
    We address the challenge of content diversity and controllability in pedestrian simulation for driving scenarios. Recent pedestrian animation frameworks have a significant limitation wherein they primarily focus on either following trajectory~\cite{rempeluo2023tracepace} or the content of the reference video~\cite{Wang_2023_ICCV}, consequently overlooking the potential diversity of human motion within such scenarios. This limitation restricts the ability to generate pedestrian behaviors that exhibit a wider range of variations and realistic motions and therefore restricts its usage to provide rich motion content for other components in the driving simulation system, \eg, suddenly changed motion to which the autonomous vehicle should respond. In our approach, we strive to surpass the limitation by showcasing diverse human motions obtained from various sources, such as generated human motions, in addition to following the given trajectory. The fundamental contribution of our framework lies in combining the motion tracking task with trajectory following, which enables the tracking of specific motion parts (\eg, upper body) while simultaneously following the given trajectory by a single policy. This way, we significantly enhance both the diversity of simulated human motion within the given scenario and the controllability of the content, including language-based control. Our framework facilitates the generation of a wide range of human motions, contributing to greater realism and adaptability in pedestrian simulations for driving scenarios. More details are in our \href{https://wangjingbo1219.github.io/papers/CVPR2024_PACER_PLUS/PACERPLUSPage.html}{project page}.
\end{abstract}    

\section{Introduction}~\label{sec:intro}
Autonomous vehicle (AV) simulation systems have gained increasing attention, given their potential to help develop safe and adaptable self-driving algorithms. One of its crucial functionalities is creating realistic and diverse pedestrian animations to train self-driving algorithms to react to a diverse array of human behaviors. It can be crucial for the safety of AV since subtle changes in pedestrians' moving directions or gestures could entail large changes in vehicle behaviors. However, despite the promising results~\cite{suo2021trafficsim,chen2021geosim,rempe2022strive} of the background scene and vehicle motion creation in the current simulation system, the performance of pedestrian animation still lags behind.

While state-of-the-art pedestrian animation methods often use keyframe animations authored by artists~\cite{drivesim}, they lack the proper reaction to the scene geometry due to the absence of the laws of physics. Recent physics simulation-based pedestrian animation method~\cite{rempeluo2023tracepace} can create pedestrian animations that are human-like, physically plausible, and conform to the geometry of the scene. Yet its animation is only controlled by 2D trajectories and is limited to basic locomotion such as walking and running, which makes it insufficient to reflect the natural diversity of pedestrian behaviors. Alternatively, pedestrian animation can also be obtained from video sequences via simulation-based motion capture~\cite{Wang_2023_ICCV}, which, however, adopts a per-video optimization strategy that is computationally intensive and thus cannot create \textit{new} and \textit{diverse} animation at a large scale or in an on-demand manner.

In this work, we propose $\name$, a simulation-based framework for generating diverse and natural pedestrian animation \textbf{on-demand}. Our framework offers richer \textit{zero-shot} control beyond trajectory following and enables the creation of diverse animation in both manual and real-world scenarios, to meet the demand for more controllable generation. Specifically, $\name$ supports fine-grained control over different body parts while following the given trajectory, which is achieved by selectively tracking \textit{specific body parts} instead of rigidly tracking the entire body~\cite{Luo2023PerpetualHC, luo2021dynamics, won2020scalable}. This creates room for more life-like animation, such as walking while making a phone call, and simultaneously ensures smoothness of the motion, compatibility of the terrain, and adherence to the provided trajectory. Using our framework, a variety of pedestrian behaviors can be introduced into the simulation system from various sources, including motion generation models, pre-captured motions, and videos, as depicted in Figure~\ref{fig:teaser}. Moreover, for the demand of recreating real-world pedestrian animation into simulation environments, $\name$ can also demonstrate motion from the given video without re-training or fine-tuning, where the missing part will be infilled automatically, as shown in Figure~\ref{fig:teaser}.

The key insight behind $\name$ lies in the synergy between motion imitation and trajectory following tasks. While the lower-body motion is often influenced by the trajectory and terrain, the upper-body motion has the flexibility to encompass a diverse range of motions. Therefore, we establish a synergistic relationship between motion imitation and trajectory following tasks through a joint training scheme. In this scheme, a single policy is employed to track partial body motion and follow trajectories simultaneously in a physically plausible way. To achieve this, we introduce a per-joint spatial-temporal mask that indicates the presence of a reference motion for the policy to track. During training, we randomly select time steps and joints to insert as the reference motion into the trajectory following task. This encourages the policy to concurrently track the trajectory and imitate the reference motion, enabling generalizable trajectory and motion tracking.

Our contributions can be summarized as follows: (1) We propose a unified physics-based pedestrian animation framework, named $\name$, which can control a simulated pedestrian to follow the 2D trajectory and \textit{specific body parts} reference motion at the same time \textit{on-demand}. (2) Our framework supports the generation of diverse pedestrian behaviors from various sources, including generative models, pre-captured motions, and videos, in any given driving scenario, such as manually built or real scanned environments. (3) Notably, our framework achieves the \textit{zero-shot} recreation of real-world pedestrian animations into simulation environments, where the missing part will be infilled automatically.
\section{Related Works}~\label{sec: related works}
\noindent\textbf{Controllable Character Animation.}
Controllable character animation has been a longstanding research topic in computer graphics and robotics~\cite{yin2007simbicon,lee2010data,muico2009contact}. Previous research in controllable character animation has often focused on integrating high-level tasks, such as trajectory following or goal-reaching, with low-level control of body joints, involving joint positions or angles. By combining these two levels of control, researchers aimed to achieve controllable animation that adheres to specific tasks or objectives. Recent methods have explored primarily two main approaches: (1) kinematics-based~\cite{holden2017phase,ling2020character, wan2023tlcontrol, zhou2023emdm} methods and (2) physics-based method~\cite{peng2021amp, peng2022ase, won2022physics, yao2022controlvae, dou2023c}. These works primarily aim to achieve predefined tasks with plausible human motions.

More recently, researchers have begun to extend the range of motion content while still adhering to given tasks. For instance, PADL~\cite{2022-SA-PADL} and CALM~\cite{tessler2023calm} introduce language-based and example-based control to generate diverse motions for the given tasks. Some recent works~\cite{xu2023composite, bae2023pmp} introduce spatial composition to expand the range of skills for more complex tasks. Based on the success of ControlNet~\cite{zhang2023adding}, AdaptNet~\cite{xu2023adaptnet} incorporates a similar design choice into its policy network to generate diverse human motions on complex terrains.

The key distinction between our work and these existing approaches lies in our focus on zero-shot fine-grained control for character animation, specifically for following given tasks. Once trained, our method does not require additional policy network training for new skills~\cite{xu2023adaptnet,xu2023composite}. Furthermore, our control framework enables flexible yet fine-grained control over the given character, including the location of upper body joints of specific examples, which has not been fully addressed in previous style-based controlling works~\cite{2022-SA-PADL,tessler2023calm,bae2023pmp}. Moreover, our approach supports motion content from various sources, such as videos, motion capture data, or even motions generated by other methods. This capability enhances the versatility and adaptability of our framework, allowing for on-demand pedestrian animation. Users can generate desired character behaviors by leveraging the flexibility of incorporating diverse motion sources.

\noindent\textbf{Physics-based Humanoid Motion Tracking.} Using a deep neural network~\cite{peng2018deepmimic,yuan2020residual,luo2021dynamics,Luo2023PerpetualHC,won2020scalable,luo2023universal} to track kinematics human motions in physics simulation achieves promising results in recent years. To achieve a better success rate, previous works introduce residual force~\cite{yuan2020residual} and Mixture-of-Experts network structures~\cite{won2020scalable,Luo2023PerpetualHC, luo2023universal}. However, unlike tracking all upper bodies, our framework allows for selective tracking of \textit{specific body parts} within the upper body and following the given trajectory.

\noindent\textbf{Physics-based Human Motion Capture.}
In recent years, the research community has developed various framework to recover human 3D poses~\cite{dabral2018learning,mehta2018single,moon2019camera,wang2020hmor,pavllo20193d,mehta2017vnect,anguelov2005scape,lassner2017unite} and motions~\cite{pavlakos2018learning,hmrKanazawa17,tan2017indirect,arnab2019exploiting,huang2017towards,sun2019human,joo2020exemplar,kolotouros2019spin,song2020human,iqbal2021kama,li2021hybrik,li2023niki, dou2023tore} from images and videos~\cite{kocabas2020vibe,rempe2021humor,yuan2022glamr}. To ameliorate the physical artifacts (\eg foot slidings) associated with the captured motion, recent work has sought to take advantage of the physical attributes of human dynamics. These methods can be broadly classified into three categories: (1) post-optimization based methods during test time~\cite{rempe2020,shimada2020physcap,xie2021physics, dabral2021gravity}, (2) reinforcement learning (RL) based methods\cite{yuan2020residual,yuan2019ego, yuan2021simpoe,peng2019mcp,luo2021dynamics,yu2021human,Luo2022EmbodiedSH,Wang_2023_ICCV} with motion imitation, and (3) physics-aware models~\cite{shimada2021physaware,zell2020weakly,li2022d} to adjust global trajectories.

Our framework is also capable of capturing physically plausible human motion via tracking high-confidence keypoints~\cite{yuan2022glamr, Wang_2023_ICCV}. However, the main objective of our paper is to achieve zero-shot reproduction of pedestrian motions in real-world driving scenarios. In contrast to existing approaches, our framework does not involve additional optimization for infilling missing frames and low-confidence motions for the captured motion in real-world driving scenarios while tracking high-confidence motions. After we reproduce these real-world scenarios, our framework is also capable of argument these environments with additional virtual pedestrians or editing infilled frames.

\section{Methodology}~\label{sec/methodology}
In this paper, we mainly focus on building up on-demand control of pedestrian animation, which encompasses two main aspects: (1) trajectory following on terrains, which determines the desired path of the simulated pedestrian in complex environments, and (2) motion content control, which specifies the desired actions and gestures exhibited by the pedestrian (\eg, making a phone call or waving a hand) while adhering to the provided trajectory and terrain.

To achieve our objective, our framework builds upon PACER~\cite{rempeluo2023tracepace} and investigates the synergy between motion imitation and trajectory following tasks. In the context of pedestrian animation in driving scenarios, the lower body motion is typically influenced by the trajectory and terrain, while the upper body motion can leverage rich semantic information specific to pedestrians. This grants the upper body the freedom to track a diverse range of possible motions. To attain fine-grained control over different body parts we introduce a per-joint spatial-temporal mask rather than tracking all body parts throughout the sequence. This mask indicates the presence of a reference motion that the policy should track. Using this tracking task, our framework enables diverse pedestrian behaviors at specific time steps and locations in a \textit{zero-shot} manner. This means that we can generate a wide range of motion behaviors without the need for additional training or optimization. Our framework also seamlessly integrates generative human motion models, motion capture sequences, and videos into the simulation system. 

Our framework is designed not only for manually synthetic scenarios but also for simulating pedestrians from real-world videos, as demonstrated in~\cite{Wang_2023_ICCV}. To enable accurate tracking of various parts of pedestrian motion in real-world videos, we expand the spatial-temporal mask to cover whole-body joints instead of solely the upper body. This enhancement allows our framework to track high-confidence motion obtained from pose estimation methods, particularly in real-world captured driving scenarios. By incorporating this capability, our framework becomes more versatile and applicable, showcasing its potential for realistic synthesis and tracking of pedestrian motion in real-world settings. This feature ensures smooth continuity and accuracy in the animation when integrating real-world data into the simulated environments while preserving the motion characteristics observed in real-world scenarios.

In Section \ref{sec:controller} to Section~\ref{sec:motion tracking}, we provide detailed insights into our controller. Subsequently, in Section \ref{sec:integration}, we delve into the integration of different motion content and scenarios within our framework. We discuss how our controller seamlessly adapts to various types of motion content, including generative models, motion capture sequences, and videos. Furthermore, we explore the applicability of our framework to different scenarios, allowing the generation of diverse pedestrian behaviors in specific contexts.

\begin{figure*}[t!]
	\centering
	\includegraphics[width=0.9\linewidth]{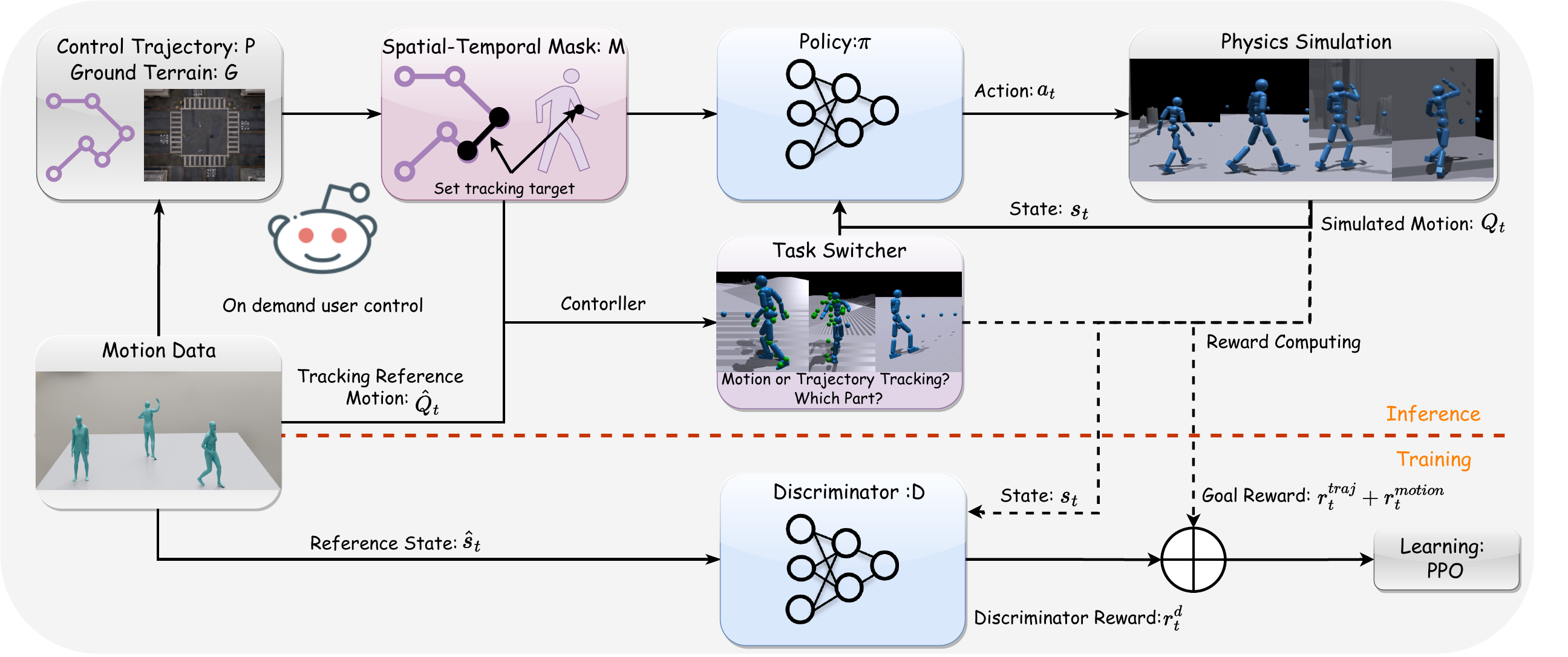}
	\caption{\small Framework of $\name$. Our framework follows the goal-conditioned reinforcement learning with Adversarial Motion Prior. To enable fine-grained control of specific body parts, we introduce an additional spatial-temporal mask to the motion-tracking task. This mask indicates the presence of a reference motion that the policy should track. By focusing on this tracking task, our framework enables the demonstration of diverse pedestrian behaviors at specific time steps and locations in a \textit{zero-shot} manner.}
    \label{fig:method}
    \vspace{-0.45cm}
\end{figure*}

\subsection{Physics-aware Character Control}~\label{sec:controller}
In this section, we first present the formulation of our pedestrian animation controller. In the following Section~\ref{sec:traj following} and Section~\ref{sec:motion tracking}, we will introduce the details of the tasks in this framework.

\paragraph{Formulation.}
We follow the general framework of goal-conditioned RL, as shown in Figure~\ref{fig:method}. The objective of our controller encompasses two aspects: (1) faithfully following the given trajectory $\mathcal{P}$ on terrain $\mathcal{G}$, and (2) imitating the specified motion content $\hat{\mathcal{Q}} = \hat{q}_{1:t}$ provided by our content module within the designated time range $\{t_s: t_s + t\}$ along the trajectory. 

Similar to prior works~\cite{peng2021amp,rempeluo2023tracepace,Luo2023PerpetualHC}, we formulate our character control as a Markov Decision Process (MDP) defined by the tuple $\mathcal{M} = \{\mathcal{S}, \mathcal{A}, \mathcal{T}, \mathcal{R}, \gamma\}$, comprising states, actions, transition dynamics, reward function, and the discount factor. The state $s_t \in \mathcal{S}$ and the transition dynamics $\mathcal{T}$ are determined by the underlying physics simulator, while the action $a_t \in \mathcal{A}$ is computed by our policy network. The reward $r_t \in R$ relates to the given trajectory and motion-tracking task. The objective of our policy is to maximize the accumulated discounted reward $\sum_{t=0}^T \gamma^t r_t$, where $\gamma$ represents the discount factor. To accomplish this, we employ the widely adopted proximal policy optimization (PPO) algorithm~\cite{schulman2017proximal}. 

\paragraph{State and Actions.} In our framework, the state $s_t \triangleq (s_t^p, s_t^g)$ consists of humanoid proprioception~\cite{Luo2023PerpetualHC} $s_t^p$ and the goal state $s_t^g$. The goal state $s_t^g$ consists of two components, as the goal for trajectory following $s^{traj}_t$, and the goal for motion tracking $s_t^{motion}$. We will present the details of these two components in the following sections. We use a proportional derivative (PD) controller at each degree of freedom (DoF) of the humanoid to control pedestrian animation.

\paragraph{Adversarial Motion Prior.}
Similar to the previous state-of-the-arts~\cite{peng2021amp, peng2022ase, rempeluo2023tracepace,Luo2023PerpetualHC}, we learn our optimal control policy with Adversarial Motion Prior (AMP). AMP employs a motion discriminator to encourage the policy to generate motions that align with the movement patterns observed in a dataset of human-recorded motion clips. Specifically,  AMP uses a discriminator to compute a style reward, which is added to the task reward: $r_t = 0.5 r_t^{amp} + 0.5(r_t^{traj} + r_t^{motion}$). We will illustrate the details of the task reward in the following sections.
\subsection{Trajectory following on terrains}~\label{sec:traj following}
\paragraph{Trajectory Following State.}
In the trajectory following task, the humanoid a local height map $\mathcal{G}$ and the trajectory $\mathcal{P}$ to follow. The 3D trajectory input is defined as $\mathcal{P}^{traj}_t = \{\hat{p}_{t}, \hat{p}_{t+\rho},\cdots, \hat{p}_{t+N\rho}\}$, where $\rho$ is the sampling rate of the trajectory, and $N$ is the number of steps in the future. $\hat{p}_{t+\rho}$ is the relative $xy$ value between the position of path $\mathcal{P}$ at time step $t + \rho$ and the root position of simulated character at time step $t$. In practice, we set $\rho$ as 0.5 seconds and $N$ as 10. For the height map of the ground terrain $\mathcal{G}^t$ , we render a 20x20 square centered at the root of the humanoid and render the local height map as input $\mathcal{G}^t$. Therefore, the goal state of the trajectory following task can be defined as $s^{traj}_t \triangleq (\mathcal{P}^{traj}_t, \mathcal{G}_t)$.

\paragraph{Trajectory Following Reward and Early Termination.}
Trajectory following task reward  is defined as  $xy$ distance between the position of trajectory $\hat{p}_{t}^{xy}$ and the root position of the simulated character $r_{t}^{xy}$ at time step $t$, formulated as $r_t^{traj} = e^{-2 ||\hat{p}_{t}^{xy} - r_{t}^{xy}||}$. To better follow the trajectory, we introduce an early termination mechanism to this task while training the policy network. Specifically, we terminate the trajectory following task if the distance between the position of trajectory $\hat{p}_{t}^{xy}$ and the root position of the simulated character $r_{t}^{xy}$ at time step $t$ is larger than a threshold $\tau$. We set $\tau$ at 0.5 meters in our experiments.    

\subsection{On-demand Motion Tracking}~\label{sec:motion tracking}

\paragraph{Masked Motion Tracking.}
In contrast to previous works~\cite{Luo2023PerpetualHC, luo2021dynamics, won2020scalable}, our motion tracking tasks deviate in that we require the policy network to track specific motion parts within a given time range while following trajectories. To facilitate this, we introduce a spatial-temporal mask to the tracking tasks, denoted as $\mathcal{M}{1:T} = \{m_{1}, m_{2}, \cdots, m_{T}\}$, where $m_{t}=\{m_{t}^1, \cdots, m_{t}^J\}$ is a set of binary masks indicating whether the motion tracking task $j$ is required at time step $t$. By employing this observation mask, we can define the state of the motion-tracking task and the reward function as follows.
\paragraph{Motion Tracking State.}
The motion content of our track task $\hat{q}_{t+1}$ for the frame $t+1$ consists of joint position $\hat{p}_{t+1}$, joint rotation $\hat{\theta}_{t+1}$, joint velocity $\hat{v}_{t+1}$, and rotation velocity $\hat{\omega}_{t+1}$, similar to the rotation-based imitation of PHC~\cite{Luo2023PerpetualHC}. In our simulation stage, we can only set the motion demonstration tasks for some specific frames, rather than tracking all frames as~\cite{Luo2023PerpetualHC}. In general, for frames without motion demonstration tasks at time step $t_1$, we directly set the mask as $0$ to indicate that motion tracking tasks are not required at these time steps. For the tracking target, we can directly set it as the same value as the state of the simulated character. For the frame $t_2$ with target motion, we can set the mask $m_{t_2}=\{m_{t_2}^1, \cdots m_{t_2}^J \}$ with $1$ for the joints that should be tracked and $0$ for the ignored joints. We also set the target motion as the same value as the state of the simulated character for the ignored joints. Therefore, the state of motion content demonstration at time step $t$ can be defined as $S_{d} \triangleq (\hat{\theta}_{t+1} - \theta_{t}, \hat{p}_{t+1} - p_{t}, \hat{v}_{t+1} - v_{t}, \hat{\omega}_{t+1} - \omega_t, \hat{\theta}_{t+1}, \hat{p}_{t+1}, m_{t+1}$).

\paragraph{Demonstration Reward and Early Termination.}
The reward of our motion demonstration is mainly related to the motion tracking error between the simulated character and the target motion. Therefore, we can define the reward as $r_t^{motion} = w_{jp}e^{-100||\hat{p}_{t} - p_{t}|| \circ m_{t}} + w_{jr}e^{-10||\hat{q}_{t} - q_{t}|| \circ m_{t}} + w_{jv}e^{-0.1 ||\hat{v}_{t} - v_{t}|| \circ m_{t}} + w_{rv}e^{-0.1 ||\hat{\omega}_{t} - \omega_{t}|| \circ m_{t}}$. The mask $m_t$ helps us to ignore the joints that should not be demonstrated in the simulation process. 

For effective training of our motion tracking task using the Adversarial Motion Primitives (AMP) approach, we made two critical design choices: incorporating additional motion sequences and implementing early termination. To address mode collapse~\cite{peng2021amp,rempeluo2023tracepace}, we trained AMP with a smaller dataset of approximately 200 sequences, as discussed previously~\cite{rempeluo2023tracepace}. While this selection ensures naturalness in generated motions, it limits generalization to unseen motion in the motion-tracking task. Including supplementary motion sequences as references in motion tracking introduces diverse motion content and has the potential to enhance tracking performance. However, training the motion tracking task with an additional dataset poses a challenge in jointly learning AMP alongside the smaller dataset as contrasting motion styles are introduced. \textit{Supplementary videos} visually depict the challenges faced by the policy network in accurately learning motion-tracking outcomes.

To overcome this limitation, we incorporate an early termination mechanism during training. Specifically, we terminate the motion demonstration task if the largest distance between the joint positions of the reference poses $\hat{p}_{t}^{xy}$ and the simulated character $p_{t}^{xy}$ at time step $t$ exceeds a threshold $\tau$. In our experiments, we set $\tau$ to 0.3 meters. We use more than 10,000 motion sequences from the AMASS dataset~\cite{mahmood2019amass} to train our motion demonstration task for practical implementation.

\subsection{System Overview}~\label{sec:integration}
\begin{figure}[t!]
	\centering
	\includegraphics[width=1.0\linewidth]{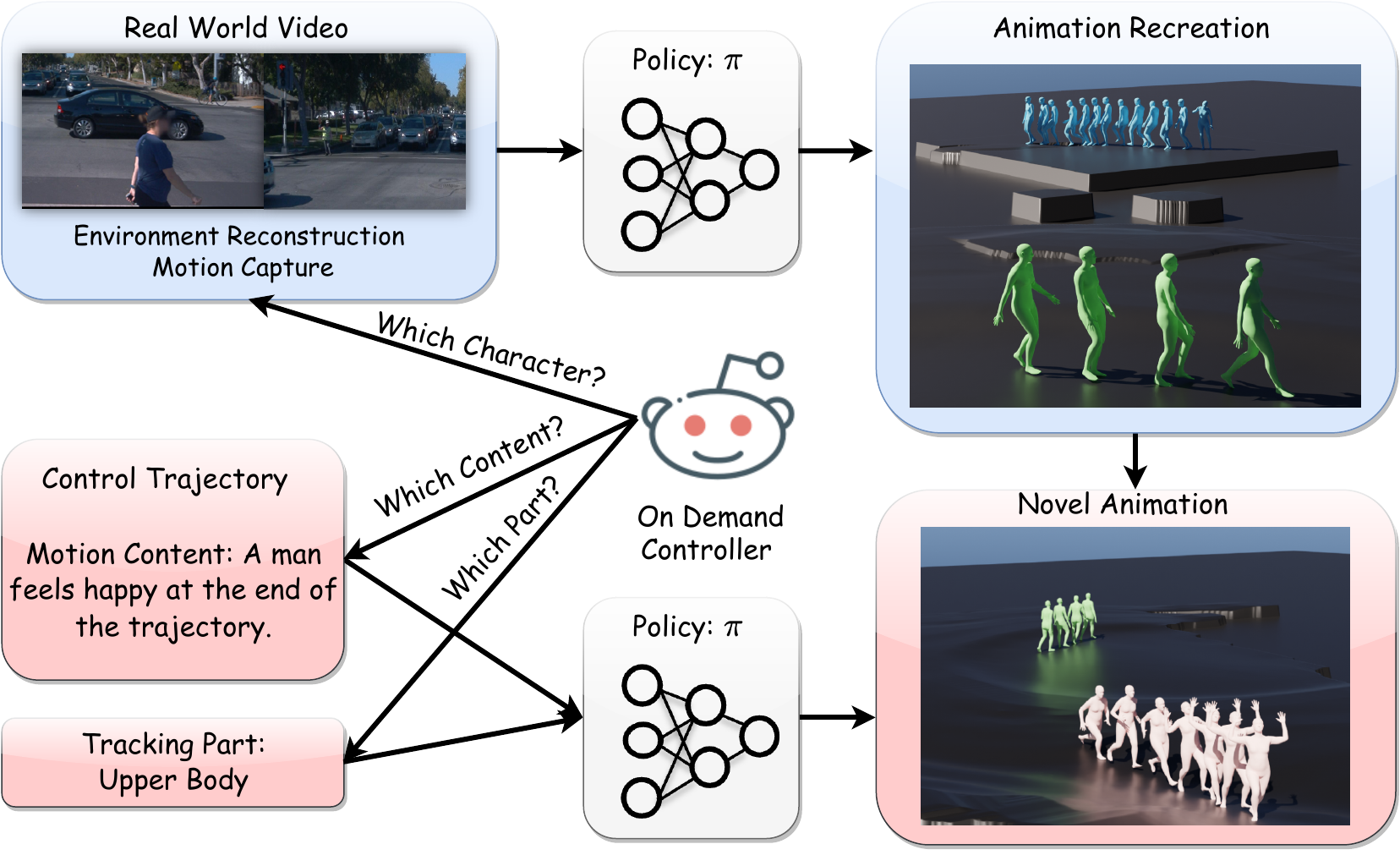}
	\caption{\small Our framework presents an on-demand control system tailored for real-world videos. Beginning with the pre-processing in~\cite{Wang_2023_ICCV}, our policy network can track high-confidence motions and effectively fill in missing parts without additional fine-tuning. Moreover, our framework offers the novel functionality of introducing customized animations into real-world scenarios with flexible control options. }
    \label{fig:usage}
\end{figure}
Finally, we outline the training process for our policy using the combination of these tasks. Subsequently, we introduce the methodology for controlling pedestrian animation on-demand using our framework in both manually synthetic and real-world scenarios.
\paragraph{Training Procedure.}
In our framework, the policy network undergoes training through a combined approach of trajectory-following and motion-tracking tasks. Initially, each training environment involves joint training of the policy with both tasks. During this step, binary masks for the reference motion are randomly generated at each time step, and early termination is applied to the motion-tracking task. The reference motions are sampled randomly from the AMASS dataset~\cite{mahmood2019amass}. Subsequently, we train the trajectory following task using randomly generated synthetic trajectories \cite{rempeluo2023tracepace,Wang_2023_ICCV}. In this stage, all joints within the spatial-temporal mask are assigned a value of 0. This ensures that the policy focuses solely on learning to follow the generated trajectories without considering motion tracking. By employing this combined training approach, we enable the policy network to acquire proficiency in trajectory-following and motion-tracking tasks, enhancing our framework's overall performance and adaptability.

\paragraph{Manually Synthetic Scenarios.} In manually synthetic scenarios, our framework offers flexibility in manually setting the trajectory while generating the desired motion content. First, we identify the specific body part from the motion content obtained from other references. Then, we set the motion tracking task's mask to indicate the desired motion for demonstration. During this process, we align the reference motion's location and orientation with the trajectory to facilitate accurate tracking. This alignment ensures that the generated motion content precisely follows the specified trajectory, allowing for diverse and customizable pedestrian animations. Our experimental results will present further details and insights on this approach. By employing this methodology, our framework empowers the generation of tailored motion content for manually synthetic scenarios, enabling greater control and realism in the animation process.

\paragraph{Real-world Scenarios.} In our real-world scenarios, we adopt the definition of high-confidence frames, as described in prior works \cite{Wang_2023_ICCV,yuan2022glamr}, using 2D keypoint detection. We track the entire body motion for these high-confidence frames to maintain optimal motion content. Conversely, in low-confidence frames, we assign a value of \textit{1} only to keypoints with high-confidence estimation scores in the spatial-temporal mask. This approach enables motion capture even when half of the body is occluded without requiring additional optimization steps. Additionally, we can apply the same process as in manually synthetic scenarios to introduce additional content from other sources into real-world scenarios using our unified policy. Figure~\ref{fig:usage} illustrates this capability. By following this approach, we enhance the quality and realism of animation in real-world videos, leveraging the flexibility of our framework.

\section{Experiments}~\label{sec:experiments} 
\vspace{-0.8cm}
\paragraph{Dataset.}
In our experiments, we utilized motion data from various sources. We employed motion from the AMASS dataset~\cite{mahmood2019amass} for motion tracking evaluation. To enhance the diversity of demonstrated motion, we collaborated with off-the-shelf language-based motion generation models~\cite{tevet2023human,chen2023executing}. Additionally, we utilized NIKI \cite{li2023niki}, a state-of-the-art human motion capture approach, to capture motions from videos and recreate real-world scenarios. Regarding the simulation environment in our framework, it encompasses two aspects: (1) manually synthetic scenarios built using Unreal Engine following the MatrixCity framework~\cite{li2023matrixcity}, and (2) real-world scenarios reconstructed from scanned point cloud data in the Waymo Open Dataset~\cite{zheng2022multi}. Following the methodology described in~\cite{Wang_2023_ICCV}, we resampled human motion captured from videos to 30 fps to match the simulation environment. To evaluate the performance of our framework effectively, we selected motion sequences with a trajectory length of more than 3 meters.

\begin{figure}[t!]
	\centering
	\includegraphics[width=1.0\linewidth]{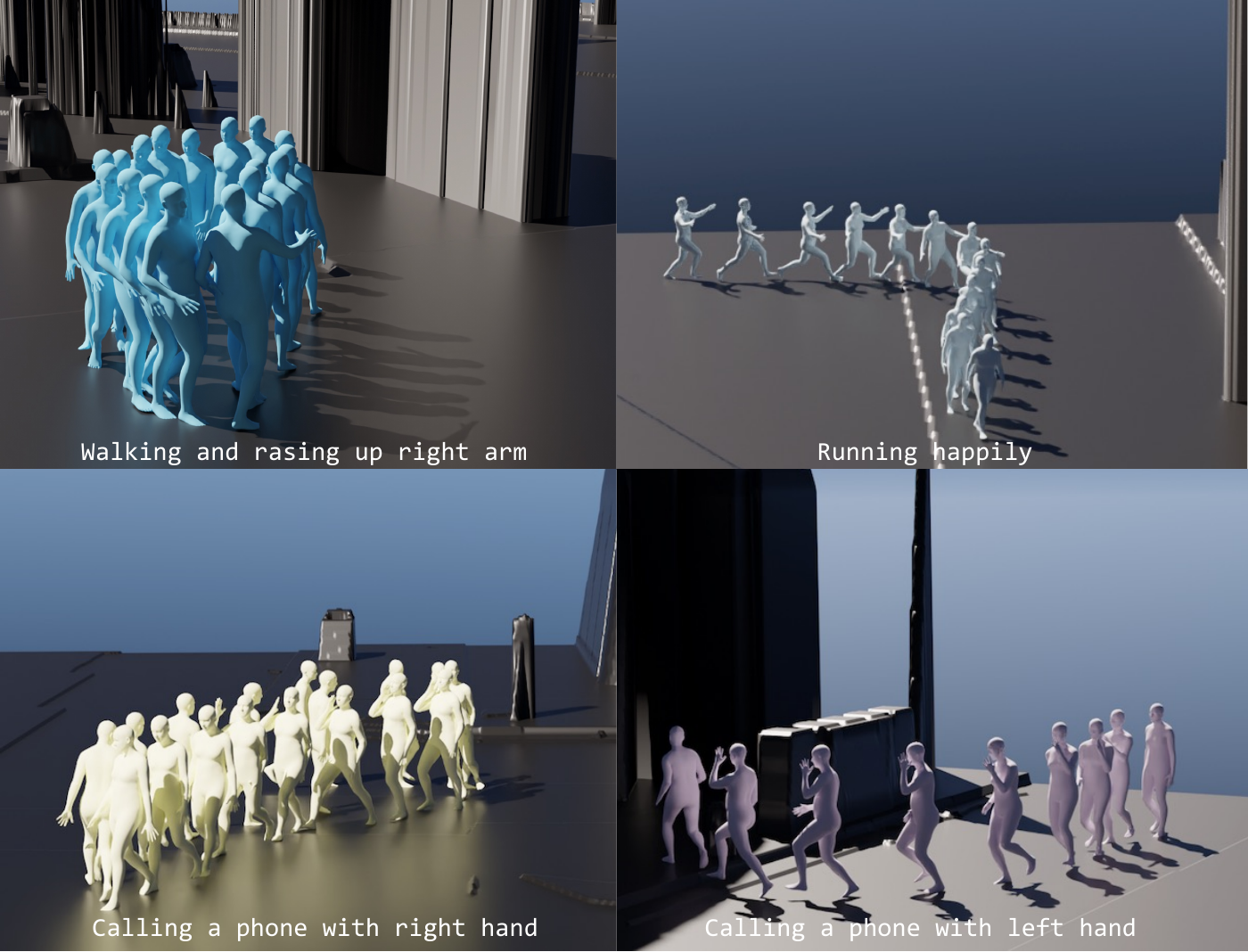}
	\caption{\small Results on manually synthetic terrains. Our framework enables the synthesis of animations by combining a given trajectory with motion content generated by language-based motion generation models~\cite{tevet2023human,chen2023executing}.}
	\label{fig:synthetic}
	\vspace{-0.5cm}
\end{figure}
\paragraph{Metrics.} To evaluate our framework, we employed a range of kinematics-based and physics-based metrics. We use motion Fréchet Inception Distance (FID)~\cite{li2020learning,petrovich2021action} and diversity metric~\cite{chen2023executing,tevet2023human} to evaluate the quality and diversity of synthesized animations. To evaluate tracking accuracy, we employed the Mean Per-Joint Position Error ($E_{mpjpe}$) and Global Mean Per-Joint Position Error ($E_{gmpjpe}$) metrics, between the simulated character and the reference motion in root space and global space. Regarding the physical attributes of the animation, we evaluated food sliding (FS) and foot penetration (FL) metrics for animation synthesis, following the methodologies outlined in~\cite{yuan2021simpoe,li2022d}. Motion jitter is computed by the velocity (Vel) and acceleration (Accel) between the physics character and the reference motion. The units for these metrics are measured in millimeters ($mm$), except for Accel, which is measured in $mm/frame^2$. 

\paragraph{Implementation Details.}
We followed the capsule model of the SMPL robot as the simulation target, as described in~\cite{Luo2023PerpetualHC,rempeluo2023tracepace,luo2023universal}. Our policy network was trained on a single NVIDIA A100 GPU, which took approximately three days to converge. Once trained, the composite policy runs at a frame rate exceeding 30 FPS. The physics simulation is performed in NVIDIA's Isaac Gym~\cite{makoviychuk2021isaac}. The control policy operates at 30 Hz, while the simulation runs at 60 Hz. In our evaluation, we did not consider body shape variation and used the mean body shape of the SMPL.
\begin{figure*}[t!]
	\centering
	\includegraphics[width=0.9\linewidth]{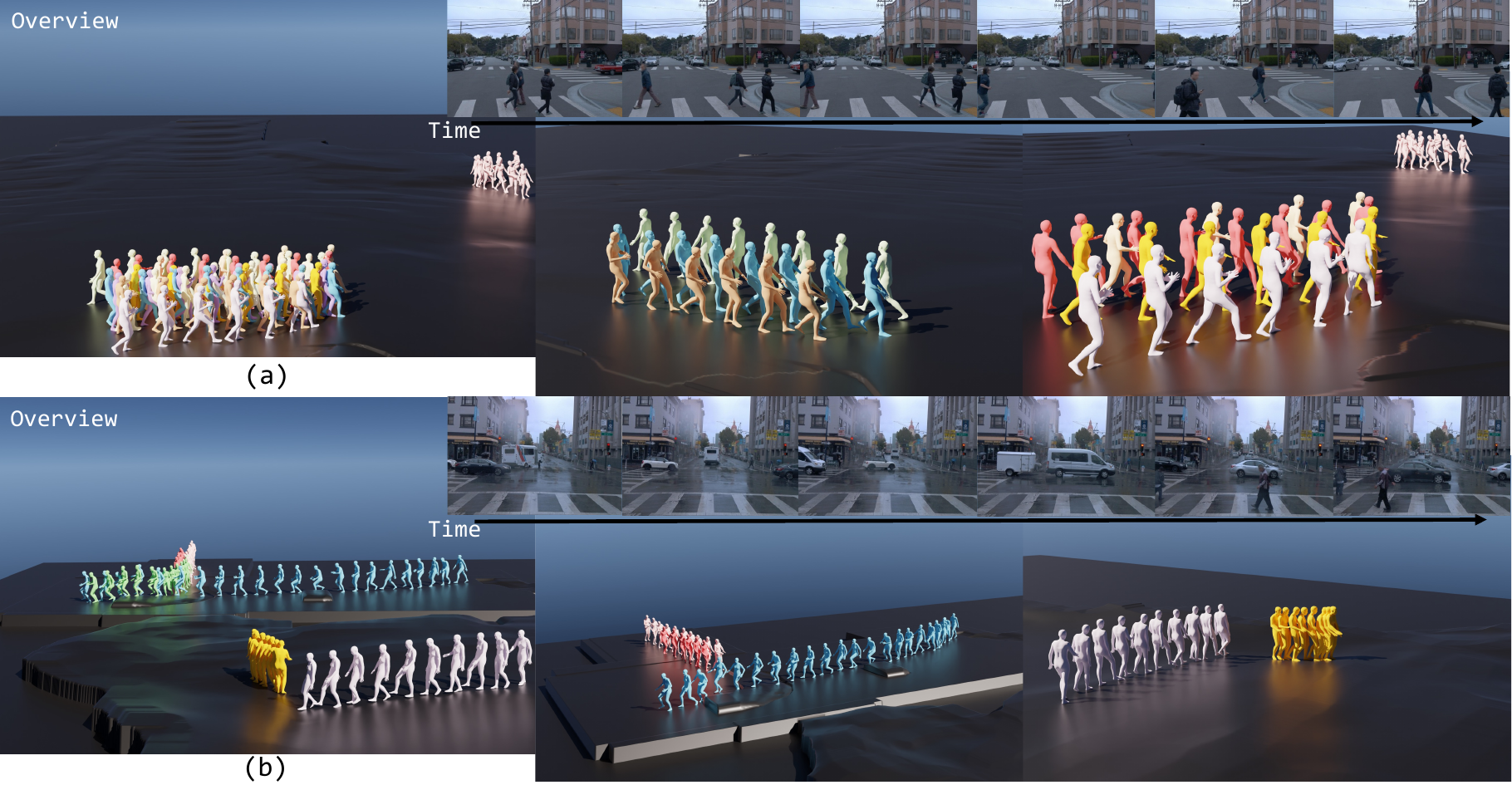}
	\caption{\small Zero-shot animation recreation of real-world pedestrians. Our framework is capable of simulating pedestrian animation following the motion content of real-world videos.}
    \label{fig:real-world}
	\vspace{-0.5cm}
\end{figure*}
\subsection{Evaluation.}
In this section, our framework is primarily compared with PACER~\cite{rempeluo2023tracepace}, the state-of-the-art controllable pedestrian animation approach. The comparison focuses on motion quality and evaluation metrics for specific tasks, such as trajectory following and motion tracking. By comparing our framework with PACER, we aim to demonstrate the advances and improvements in these areas.

\vspace{-0.3cm}
\paragraph{Motion Quality and Diversity.}
\begin{table}[t]
	\centering
	\caption{Comparison of Motion Quality and Diversity between Our framework and PACER. FID and Diversity metrics were used for trajectories with normal speed, while \textit{l}-FID and \textit{l}-Diversity are employed for animations under low speed.}\label{tab:quality-diversity}
	{\small\begin{tabular}{l |c c | c  c}
			\shline
			Method  & FID $\downarrow$ & Diversity $\uparrow$ & \textit{l}-FID $\downarrow$ & \textit{l}-Diversity $\uparrow$ \\
            \hline
            PACER~\cite{rempeluo2023tracepace} & 7.97 & 1.29 & 8.84 & 1.24\\
            \textbf{Ours} & \textbf{6.74} & \textbf{1.67} & \textbf{7.62}  & \textbf{1.36}\\
			\shline
	\end{tabular}}
\end{table}
We conducted a comparative analysis between our framework and PACER, focusing on motion quality and diversity. For this evaluation, we randomly synthesized 1000 different trajectories on the synthetic terrain, which were used to train both PACER and our policy. The motion content of our framework is synthesized by off-the-shelf approaches~\cite{tevet2023human,chen2023executing}. Table~\ref{tab:quality-diversity} presents the results of this comparison. Our method achieves a lower FID and demonstrates better diversity compared to PACER. These findings indicate the superior ability of our framework to generate diverse and contextually relevant pedestrian animations. Additionally, we consider the issue of synthesizing animations at low speeds, which can often result in unnatural motion, as presented in PACER. Specifically, we compare our framework with PACER under trajectories with low speeds (speed $<1m/s$). As shown in Table~\ref{tab:quality-diversity}, our framework consistently achieves better motion realism and diversity in these low-speed scenarios. Overall, our framework surpasses PACER in motion quality and diversity, showcasing its advancements and improvements in realistic and diverse pedestrian animation synthesis.
\paragraph{Motion Tracking.}
\begin{table}[t]
	\centering
	\caption{Motion tracking quality of our method between different body parts by introducing spatial-temporal mask to the corresponding region. We compare with~\cite{Wang_2023_ICCV} for whole-body tracking because this method can not only track specific regions, \eg, upper-body.}\label{tab:tracking}
	\scalebox{0.85}{\small \begin{tabular}{l |c | c c c c}
			\shline
			Metric & Wang~\cite{Wang_2023_ICCV} & Whole & Upper  & Left Arm  & Right Arm  \\
            \hline
            $E_{mpjpe}$ $\downarrow$  & 80.29 & \textbf{72.10} & 77.87  & 78.75 & 79.52\\
			$E_{gmpjpe}$ $\downarrow$ & 137.48 & \textbf{123.88} & 128.15 & 128.84 & 133.92\\
			\shline
	\end{tabular}}
	\vspace{-0.4cm}
\end{table}
To evaluate the motion tracking performance of our framework, we utilize motion content from two sources: the AMASS dataset \cite{mahmood2019amass} and synthesized motion generated by state-of-the-art motion generation models~\cite{tevet2023human,chen2023executing}. We randomly select 1000 sequences from AMASS to assess the tracking performance, providing diverse and real-world motion content. Additionally, we generate 200 synthesized motion sequences using ChatGPT~\cite{chatgpt} with 20 distinct prompts for driving scenarios, resulting in 10 sequences per prompt. The evaluation focuses on synthetic terrains and trajectories, with a comprehensive assessment of whole-body, upper-body, and left/right arm tracking. The tracking results are presented in Table~\ref{tab:tracking}, allowing us to analyze and quantify the effectiveness of our framework in different tracking scenarios and body parts. Furthermore, we compare our method with~\cite{Wang_2023_ICCV} for whole-body tracking, demonstrating superior \textit{zero-shot} tracking results on terrains.
\subsection{Results on Real-world Scenarios}
\begin{table}[t]
	\centering
	\caption{We present the results of our method on real-world scenarios and compare with~\cite{Wang_2023_ICCV}}\label{tab:real-world}
	\scalebox{0.8}{\small\begin{tabular}{l |c c|  c c c c}
			\shline
			Method  & $E_{mpjpe}$ $\downarrow$ &  $E_{gmpjpe}$ $\downarrow$ & FS $\downarrow$& FL $\downarrow$ & Vel $\downarrow$ & Acc $\downarrow$\\
            \hline
            Motion~\cite{li2023niki} & $\times$ & $\times$& 45.32 & 54.21  &$\times$ & $\times$\\
			\hline
            Wang~\cite{Wang_2023_ICCV} & 89.42 & 137.84 & 7.87 & 14.21& 8.21 & 7.42\\
            \textbf{Ours} & \textbf{77.67} & \textbf{127.84} & \textbf{7.68} & \textbf{12.12} & \textbf{7.42} & \textbf{6.43}  \\
			\shline
	\end{tabular}}
	\vspace{-0.4cm}
\end{table}
In real-world scenarios, we evaluate our framework using the NIKI~\cite{li2023niki} to obtain joint rotations of the human body. Following the evaluation methodology outlined in~\cite{Wang_2023_ICCV}, we use the ground truth trajectory and 2D bounding box to assess our framework's performance. To evaluate the confidence of the estimated results, we employ ViTPose~\cite{xu2022vitpose} to extract confidence scores for each body joint. During the inference process, we selectively track body parts with high-confidence joint estimations, ensuring a fair comparison by refraining from additional fine-tuning or optimization, as stated in~\cite{Wang_2023_ICCV}. This unbiased evaluation allows for a comparison of our framework's performance. Our method demonstrates improvements in the physics attributes of the motion content, as presented in Table~\ref{tab:real-world}. Moreover, it achieves better $E_{mpjpe}$ and $E_{gmpjpe}$ results, indicating improved matching to high-confidence parts and the given trajectory compared to~\cite{Wang_2023_ICCV}. Through these evaluation techniques, we showcase the results of our framework in real-world scenarios, highlighting its performance and effectiveness in practical settings.
\subsection{Ablation Study}
\begin{table}[t]
	\centering
	\caption{ Ablation studies on motion tracking and spatial-temporal mask. Our design choice achieves better results on both motion tracking and motion quality.}\label{tab:ablation}
	{\small\begin{tabular}{c c | c c c}
			\shline
			Tracking  & Mask & FID& $E_{mpjpe}$  $\downarrow$ &  $E_{gmpjpe}$ $\downarrow$ \\
			\hline
			$\times$ & $\times$ & 7.97 & 215.55 & 254.17 \\
			$\surd$ & $\times$ & 7.07 & 79.57 & 132.24 \\
			\hline
			$\surd$ & $\surd$ & \textbf{6.74} & \textbf{77.87}  & \textbf{128.15}\\

			\shline
	\end{tabular}}
	\vspace{-0.5cm}

\end{table}
We performed our ablation study at Table~\ref{tab:ablation} to assess the effectiveness of motion tracking and the spatial-temporal mask in our framework. The study focused on upper body tracking and trajectory following. When motion tracking is not included, our framework resembles PACER and can not follow the content of the given motion sequences. Consequently, the motion quality was inferior to our framework. However, upon introducing the motion tracking task, combined with the spatial-temporal mask, the policy exhibited improved motion tracking and enhanced realism quality. The results of the ablation study highlight the significance of motion tracking and the spatial-temporal mask, underscoring their contributions to the effectiveness and quality of our framework.
\subsection{Qualitative Results}
Figure \ref{fig:synthetic} showcases the synthesized animations on artificial terrains. All presented results adhere to the control of the given trajectory and upper body motion content. Our framework enables the synthesis of diverse and natural human animations, surpassing the limitations of conventional walking and running actions~\cite{rempeluo2023tracepace}. Furthermore, Figure~\ref{fig:real-world} illustrates the \textit{zero-shot} results of animation recreation. These examples highlight the capability of our framework to recreate animations in real-world scenarios. We refer viewers to our supplementary video for a more comprehensive presentation, including different tracking parts and collaborations with various motion sources. The synthesized animations on synthetic terrains and the animation recreation results demonstrate the effectiveness and versatility of our framework in generating diverse and natural human animations.

\vspace{-0.15cm}
\section{Conclusion and Limitation}~\label{sec:conclusion}
\vspace{-1.cm}
\paragraph{Conclusion:} In this paper, we introduce a novel framework for on-demand synthesis of diverse and natural pedestrian animation in driving scenarios. Our framework surpasses traditional trajectory control methods by enabling zero-shot generation of diverse motion using a range of motion content sources. To achieve this, we propose a joint tracking framework where a single policy is trained to simultaneously track the trajectory and imitate selected joints, such as upper-body joints. During training, we incorporate a spatial-temporal mask to guide the policy network in tracking specific joints within a designated time range. Our framework empowers comprehensive control over pedestrian animation in both manual and synthetic scenarios, offering a versatile tool for animation generation.
\vspace{-0.15cm}
\paragraph{Limitations and Future Works:}
Our current approach uses pre-trained motion generation models for motion content and relies on user-provided trajectories, without explicitly considering the semantic relationship between pedestrians and the environment. In future work, we aim to investigate generating motion content directly through the policy network while incorporating semantic guidance.
\vspace{-0.15cm}
\paragraph{Acknowledgement}
This work is funded in part by the National Key R$\&$D Program of China (2022ZD0160201), and Shanghai Artificial Intelligence Laboratory.
\section{Supplementary}
\subsection{Implementation Details}
\paragraph{Humanoid State.} In our policy, the state of the humanoid, denoted as $s_t^p$, comprises joint positions $j_t \in \mathbb{R}^{24 \times 3}$, rotations $q_t \in \mathbb{R}^{24 \times 6}$, linear velocities $v_t \in \mathbb{R}^{24 \times 3}$, and angular velocities $\omega_t \in \mathbb{R}^{24 \times 3}$. These components are normalized with respect to the agent's heading and root position in our simulator. The rotation $q_t$ is represented using the 6-degree-of-freedom rotation representation \cite{zhou2019continuity}.
\begin{figure}[h]
	\vspace{-0.2cm}
	\centering
	\includegraphics[width=0.9\linewidth]{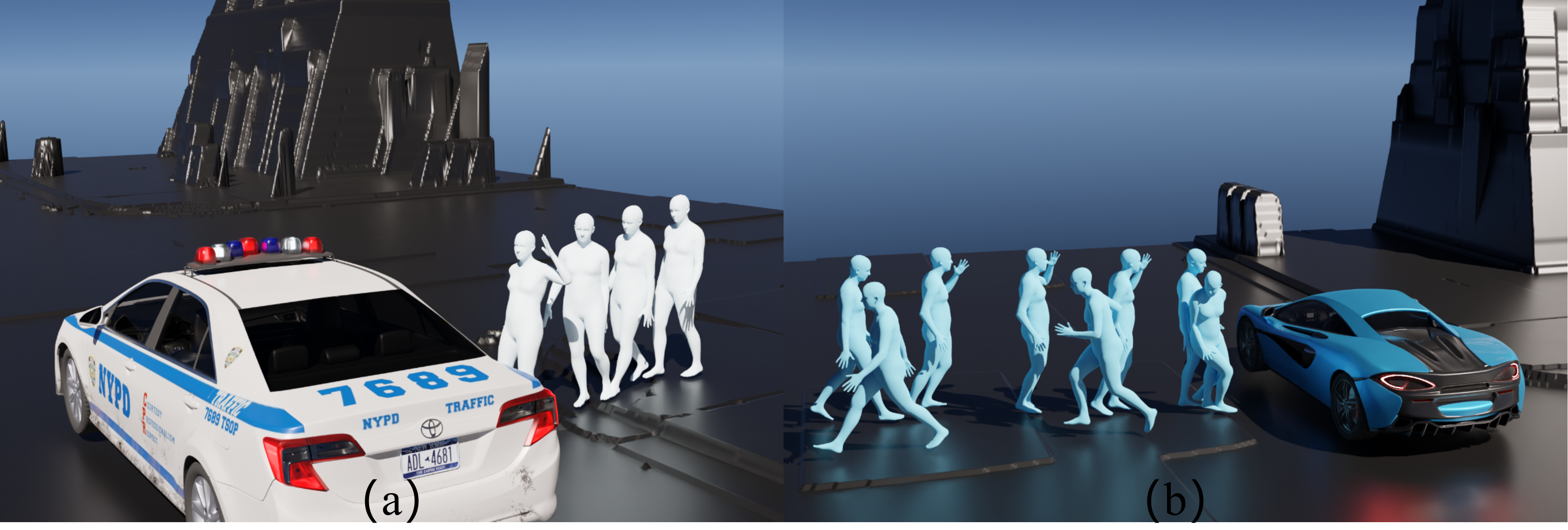}
	\caption{\small Our simulated motion near a moving car allows for interactive actions, such as waving, when the pedestrian stops or turns around. This enhances the realism of the simulated motion.}
	\label{fig:results1}
	\vspace{-0.5cm}
\end{figure}
\paragraph{Network Architecture.} In this study, we adopt the network structure from PACER \cite{rempeluo2023tracepace}, which separates the policy network into a task feature processor and an action encoder. For the task processor, we employ a convolutional neural network (CNN) to process the terrain map, following the approach outlined in \cite{rempeluo2023tracepace}. Additionally, we utilize an MLP network to encode the trajectory, reference motion, and tracking mask. Subsequently, we concatenate the humanoid state with the output of the task processor to form the input of the action network. The action network consists of MLP layers with ReLU activations, comprising two layers with 2048 and 1024 units, respectively. The output of this action network, indicated as $a_t \in \mathbb{R}^{23 \times 3}$, corresponds to the PD target of the joints on the SMPL \cite{loper2015smpl} human body, excluding the root joint.
\subsection{Additional Results.}
\begin{figure*}
	\centering
	\includegraphics[width=1.0\linewidth]{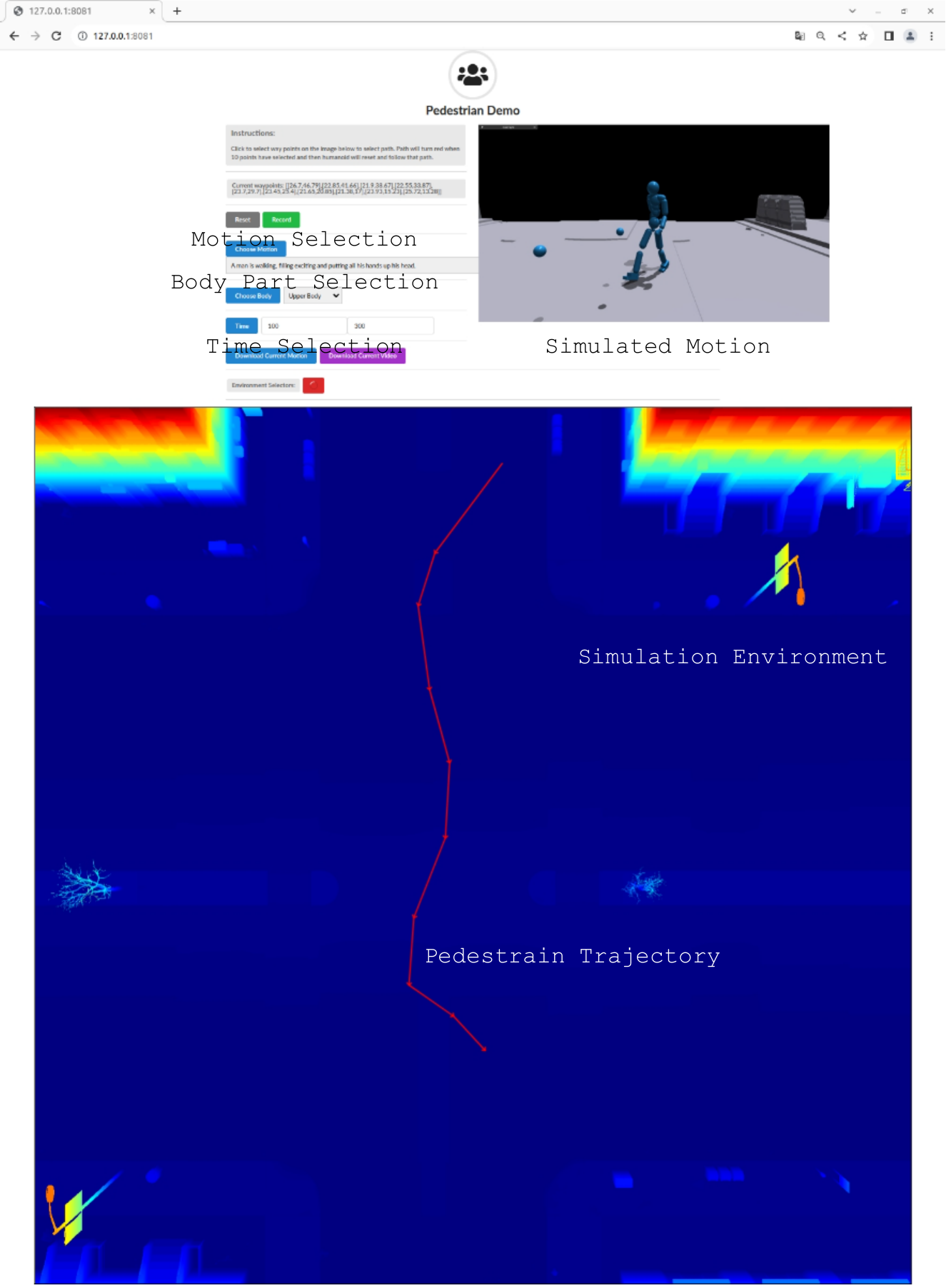}
	\caption{\small User interface (UI) of our on-demand animation controller. Our system is capable of changing trajectory, environment, motion content, and body parts for pedestrian animation in driving scenarios.}
    \label{fig:ui}
	\vspace{-0.5cm}
\end{figure*}

The trajectory following results are presented in Table~\ref{tab:supp}. We mainly compare our method with the approaches proposed in \cite{rempeluo2023tracepace} and \cite{Wang_2023_ICCV}. We evaluate these different methods on the synthetic terrain and trajectories similar to training these policies. The metric is the average deviation of the character from trajectories. As depicted in the table, PACER achieves the most favorable results. Our method demonstrates performance comparable to that of PACER in terms of trajectory following. However, it is worth noting that the kinematics policy employed in \cite{Wang_2023_ICCV} exhibits limitations when dealing with complex terrains without fine-tuning.

Additionally, as shown in Figure~\ref{fig:results1}, when dealing with a simulated car, we can manually determine the trajectory and movements of the pedestrian to enable more grounded reactions. For example, we can incorporate actions such as waving in the car or altering the direction of movement. In this work, we focus mainly on the "on-demand" prospect on increasing the capabilities of simulators, which can pave the way for a more antonymous response in intelligent agents. 

\subsection{System Details.}
Figure~\ref{fig:ui} showcases the user interface (UI) we have developed, which allows users to control the animation within the specified driving scenario. Through this system, users have the capability to modify the trajectory, motion content, body parts, and the starting time of the motion content seamlessly in a zero-shot manner. Further details regarding this system can be found in our supplementary video.
\begin{table}[t]
	\centering
	\caption{Comparison of trajectory following task with PACER~\cite{rempeluo2023tracepace} and~\cite{Wang_2023_ICCV}. Our method achieves a comparable result with PACER and a significantly better result than~\cite{Wang_2023_ICCV}.}\label{tab:supp}
	{\small\begin{tabular}{l |c c c  }
			\shline
			Method  & PACER~\cite{rempeluo2023tracepace} & Wang et.al~\cite{Wang_2023_ICCV} & Ours \\
            \hline
             Error $\downarrow$ & \textbf{0.118} & 0.164 & 0.123 \\
            \shline
	\end{tabular}}
\end{table}

\subsection{Further Discussion.}
In the realm of integrating language-based motion generation models with physics simulators, recent work has emerged to address this area. Physdiff \cite{yuan2023physdiff} stands as a notable contribution, employing a whole-body imitator within the motion diffusion model's denoising process to achieve physically plausible animations. On the other hand, InsActor~\cite{ren2023insactor} and MoConVQ~\cite{yao2023moconvq} rely on model-based character simulation. However, these approaches overlook the influence of terrain in pedestrian animation and lack body part control. Consequently, these methods are limited to controlling the entire body joints based on language instructions solely on flat-ground scenarios. In contrast, our framework offers enhanced flexibility in control while also facilitating cooperation with real-world motions on various terrains. Additionally, with the advances in motion content synthesis by language-based motion models, our framework has the potential to generate superior animations in driving scenarios.
{
    \small
    \bibliographystyle{ieeenat_fullname}
    \bibliography{main}
}


\end{document}